\documentclass[10pt,twocolumn,letterpaper]{article}

\usepackage{iccv}
\usepackage{times}
\usepackage{epsfig}
\usepackage{graphicx}
\usepackage{amsmath}
\usepackage{amssymb}
\usepackage[accsupp]{axessibility}
\def\methodname{DHANet\xspace}

\graphicspath{{figures/}{pictures/}{image/}{./}} 
\usepackage[pagebackref=true,breaklinks=true,colorlinks,bookmarks=false]{hyperref}
\usepackage[capitalise]{cleveref}

\iccvfinalcopy 

\def\iccvPaperID{1853} 

\ificcvfinal\fi

\begin{document}

\title{Dynamic Hyperbolic Attention Network for Fine Hand-object Reconstruction}

\author{Zhiying Leng$^{1,2}$, Shun-Cheng Wu$^{2}$, Mahdi Saleh$^{2}$, Antonio Montanaro$^{3}$, Hao Yu$^{2}$, Yin Wang$^{1}$,\\ Nassir Navab$^{2}$, Xiaohui Liang$^{1,4}$\thanks{corresponding author} , Federico Tombari$^{2}$\\
\small $^{1}$ State Key Laboratory of Virtual Reality Technology and Systems, Beihang University, China\\
\small $^{2}$ Computer Aided Medical Procedures, Technical University of Munich, Germany\\
\small $^{3}$ Politecnico di Torino, Italy \\
\small $^{4}$ Zhongguancun Laboratory, Beijing, China \\
{\tt\small \{zhiyingleng,liang\_xiaohui\}@buaa.edu.cn, \{shuncheng.wu,m.saleh\}@tum.de, tombari@in.tum.de}
}

\maketitle
\ificcvfinal\thispagestyle{empty}\fi

\begin{abstract}
Reconstructing both objects and hands in 3D from a single RGB image is complex. Existing methods rely on manually defined hand-object constraints in Euclidean space, leading to suboptimal feature learning. Compared with Euclidean space, hyperbolic space better preserves the geometric properties of meshes thanks to its exponentially-growing space distance, which amplifies the differences between the features based on similarity. In this work, we propose the first precise hand-object reconstruction method in hyperbolic space, namely \textbf{D}ynamic \textbf{H}yperbolic \textbf{A}ttention \textbf{Net}work (\textbf{\methodname}), which leverages intrinsic properties of hyperbolic space to learn representative features. Our method that projects mesh and image features into a unified hyperbolic space includes two modules, \ie dynamic hyperbolic graph convolution and image-attention hyperbolic graph convolution. With these two modules, our method learns mesh features with rich geometry-image multi-modal information and models better hand-object interaction. Our method provides a promising alternative for fine hand-object reconstruction in hyperbolic space. Extensive experiments on three public datasets demonstrate that our method outperforms most state-of-the-art methods.
\end{abstract}


\begin{figure}[tb]
\centering
\includegraphics[width=0.9\linewidth]{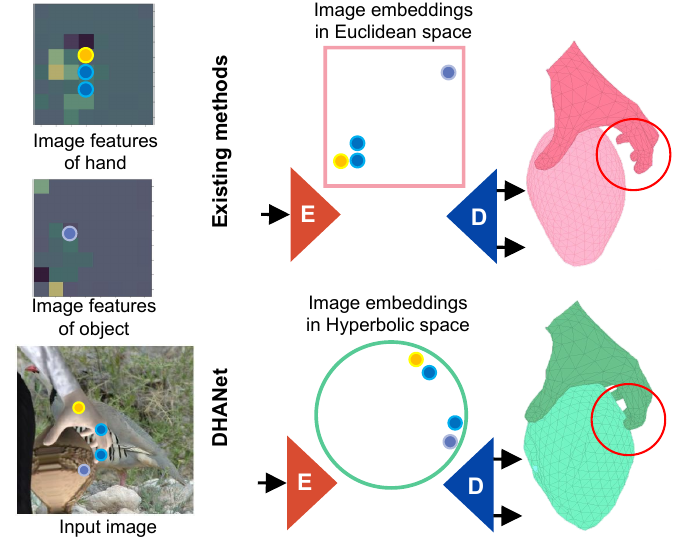}
\caption{
Colored dots indicate the features of the hand and the object. Existing methods learn image features in Euclidean space but struggle to model the contact region, resulting in separated colored dots. Our DHANet learns mesh and image features in hyperbolic space, better modeling the hand-object interaction that ensures the distribution of colored dots preserves the hand-object contact relationship. And our method results in a more accurate reconstruction, highlighted by the red circles.
}
\label{fig:teaser}       
\end{figure}%
\section{Introduction}
 3D hand-object reconstruction from monocular RGB images is a fundamental task in computer vision. Given a single RGB image of a hand interacting with an object, it aims at predicting a 3D mesh of both the hand and the object under the correct pose and precisely modeling the hand-object interaction. Although the 3D posed reconstruction has a wide application in human-machine interaction, robotic grasping/learning, and augmented reality, the challenges of this task still remain in two aspects: 1) Reconstructing meshes with the pose and scale consistent with the input; 2) Fulfilling the physiological rules on hands and physical characteristics of hand-object interaction.
%

Existing methods deal with hand-object images or meshes in Euclidean space~\cite{hasson2019learning,liu2021semi,huang2020hot, hampali2022keypoint,doosti2020hope,zhuang2021joint,tse2022collaborative,yang2021cpf,cao2021reconstructing,hampali2020honnotate}, which learn image features and regress model parameters of hand and object from Euclidean embeddings. To accurately reconstruct meshes of hands and objects, especially around the area of mutual occlusion, existing methods~\cite{liu2021semi,huang2020hot, hampali2022keypoint,tse2022collaborative,doosti2020hope,zhuang2021joint} optimize the reconstruction by taking the physical interaction between the hand and the object as a cue. These methods can be broadly divided into two categories: learning-based methods and optimization-based methods. Learning-based methods employ attention mechanism~\cite{liu2021semi,huang2020hot, hampali2022keypoint}, and other advanced models~\cite{tse2022collaborative,doosti2020hope,zhuang2021joint} to model hand-object interactions. Optimization-based methods integrate physical constraints, like Spring-mass System~\cite{yang2021cpf} and 3D contact priors ~\cite{cao2021reconstructing,hampali2020honnotate} with contact loss functions, to constraint the optimization process. Existing methods almost directly regress the model parameters of hand-object meshes from image features and manually define interaction constraints without exploiting the geometrical information. In this work, we seek for learning geometry-image multi-modal features in hyperbolic space to reconstruct accurate meshes.

As mentioned in recent research on Representation Learning in hyperbolic space~\cite{hyperbolicembedding,hyperbolicimage,hyperbolicpointcloud,peng2021hyperbolic,yang2022hyperbolic}, the effectiveness of Euclidean space for graph-related learning tasks is still bounded, failed to provide powerful geometrical representations. Compared to Euclidean space, hyperbolic space exhibits the potential to learn representative features. Due to the exponential growth property of hyperbolic space, it is innately suitable to embed tree-like or hierarchical structures with low distortion while preserving local and geometric information~\cite{peng2021hyperbolic,yang2022hyperbolic}. There have been attempts to represent and process mesh and image features in hyperbolic space~\cite{jin2006computing,shi2013hyperbolic,shi2016shape,aigerman2016hyperbolic,hyperbolicimage,hyperbolicembedding}. 
However, joint feature learning of meshes and images in hyperbolic space for accurate hand-object reconstruction has not yet been explored.

To this end, we propose the first method based on hyperbolic space for hand-object reconstruction, named Dynamic Hyperbolic Attention Network (\methodname), to leverage the benefits of hyperbolic space for geometrical feature learning (see \cref{fig:teaser}). Our approach consists of three modules, image-to-mesh estimation, dynamic hyperbolic graph convolution, and image-attention hyperbolic graph convolution. Firstly, the image-to-mesh estimation module geometrically approximates the hand and object from an input image. Secondly, hand and object meshes are projected to hyperbolic space for better preserving the geometrical information. Our dynamic hyperbolic graph convolution dynamically builds neighborhood graphs in hyperbolic space to learn mesh features with rich geometric information. Thirdly, we project mesh and image features to a unified hyperbolic space, preserving the spatial distribution between hand and object. Our image-attention hyperbolic graph convolution embeds the distribution into feature learning and models the hand-object interaction in a learnable way. With these modules, our method learns more representative geometry-image multi-modal features for accurate hand object reconstruction. Comprehensive evaluations of our method on three public hand-object datasets, namely Obman dataset~\cite{hasson2019learning}, FHB dataset~\cite{FHB}, and HO-3d dataset~\cite{hampali2020honnotate}, where DHANet outperforms most state-of-the-art methods, confirm the superiority of our design.

The main contributions of our work are as follows: 
\begin{itemize}
    \item We are the first to address hand-object reconstruction in hyperbolic space, proposing a novel Dynamic Hyperbolic Attention Network. 
    \item We devise a Dynamic Hyperbolic Graph Convolution to dynamically learn mesh features with rich geometry information in hyperbolic space.
    \item We introduce an Image-attention Graph Convolution to learn geometry-image multi-modal features and to model hand-object interactions in hyperbolic space.
\end{itemize}

\section{Related work}

\subsection{Hand-Object Reconstruction}
Hand-object reconstruction is an attractive research area. Earlier methods focused on reconstructing hand and object from multi-view images~\cite{ballan2012motion,oikonomidis2011full} or RGBD images~\cite{hamer2010object,hu2022physical} due to severe occlusion between hand and object. In recent trends, joint reconstruction of both shapes from a single RGB image has become popular. It is a more challenging task due to the limited perspective. Existing methods can be divided into two categories: optimization-based and learning-based methods. 

\textbf{Optimization-based methods} design contact patterns manually based on a parameterized representation of hand and object to model the hand-object interaction explicitly. Cao \etal.~\cite{cao2021reconstructing} leveraged the 2D image cues and 3D contact priors to constrain the optimizations. 2D image cues include the estimated object mask via differentiable rendering and the estimated depth. 3D contact priors are based on hand-object distance and collision. Yang \etal.~\cite{yang2021cpf} presented an explicit contact representation, Contact Potential Field (CPF). Each contacting hand-object vertex pair is treated as a spring-mass system. They also introduced contact constraint items and grasping energy items in their learning-fitting hybrid framework. Ye \etal.~\cite{ye2022s} parameterizes the object by signed distance, leveraging the input visual feature and output hand mesh information to infer the object representation. Zhao \etal.~\cite{zhao2022stability} represents hand and object as a hand-object ellipsoid, recovering hand-object driven by the simulated stability criteria in the physics engine. However, the performance of these methods is limited by the manually defined interaction. 

\begin{figure*}[tb]
\centering
\includegraphics[width=0.95\linewidth]{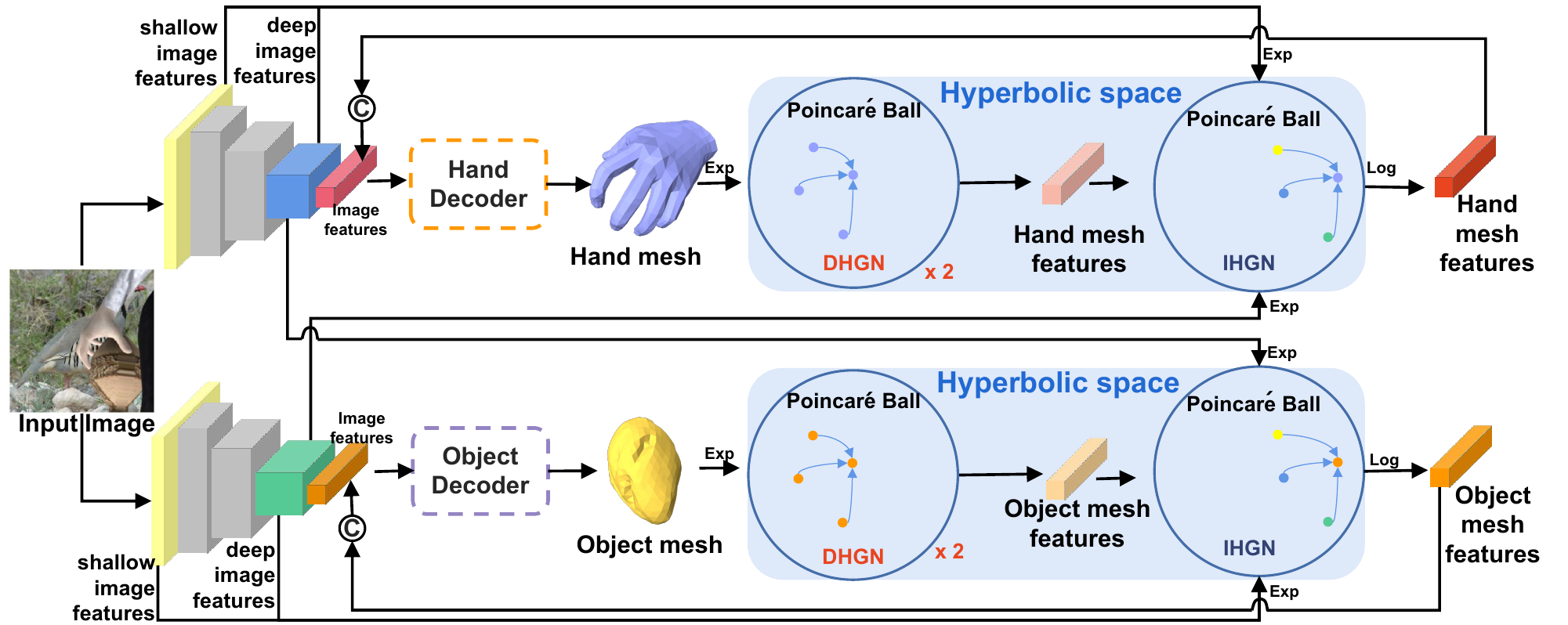}
\caption{DHANet overview. Given an image with hand-object interaction, image encoder-decoders first approximate the mesh with an initial form. Subsequently, image features from encoders and meshes are projected to hyperbolic space via the $Exp$ function. Our dynamic hyperbolic graph convolution (DHGN) and image-attention hyperbolic graph convolution (IHGN) learn representative mesh features, projected to Euclidean space via the $Log$ function and concatenated with image features to derive an accurate hand-object reconstruction.}
\label{fig:pipeline}       
\end{figure*}

\textbf{Learning-based methods} employ advanced mechanisms to model the relationship between hand and object implicitly. These methods can be divided into two categories: non-graph-based and graph-based. \textbf{Non-graph-based methods} model without the use of graph structure.
The first end-to-end learnable model is presented by Hasson \etal.~\cite{hasson2019learning} that exploits a contact loss to model the interaction. Cheng \etal.~\cite{cheng2021semi} propose a pose dictionary learning module to distinguish infeasible poses. Liu \etal.~\cite{liu2021semi} builds a joint learning framework where they performed contextual reasoning between hand and object representations. Li \etal.~\cite{yang2022artiboost} propose ArtiBoost, a lightweight online data enhancement method that constructed diverse hand-object interactions using a data enhancement approach. \textbf{Graph-based methods} represents hand-object as graphs, utilizing graph convolution to learn the hand-object interaction. Doosti \etal.~\cite{doosti2020hope} are the first to design an Adaptive Graph U-Net to transform 2D keypoints to 3D. A context-aware graph network and a learnable physical affinity loss are proposed to learn interaction messages \cite{zhuang2021joint}. Tse \etal.~\cite{tse2022collaborative} transfer mesh information to the decoder of image features in a collaborative learning strategy. An attention-guided graph convolution learns mesh information. However, these methods learn the embedding of keypoints or meshes in Euclidean space, failing to capture rich geometry information. In our work, we aim to capture geometry information in hyperbolic space,
which are beneficial to the reconstruction of hands and objects. %

\subsection{Hyperbolic Neural Networks}
\label{sec:related gnn}
Recently, incremental works have been done for deep representation learning in hyperbolic spaces~\cite{peng2021hyperbolic}. Compared with Euclidean space, hyperbolic space is more suitable for processing data with a tree-like structure or power-law distribution, owing to its exponential growth property~\cite{yang2022hyperbolic}. The deep representation learning in hyperbolic space is named hyperbolic neural networks. Existing research mostly focuses on NLP tasks~\cite{shimizu2020hyperbolic,gulcehre2018hyperbolic,zhu2020hypertext,tay2018hyperbolic}. Recently, some researchers propose to learn hyperbolic embeddings for images in computer vision tasks~\cite{khrulkov2020hyperbolic,peng2020mix,atigh2022hyperbolic}. It has been confirmed that a similar hierarchical structure exists in images as well. Montanaro \etal.~\cite{montanaro2022rethinking} is the first to apply hyperbolic neural networks to point clouds, demonstrating that a point cloud is a local-whole hierarchical structure. As far as we know, we are the first to seek hand-object reconstruction in hyperbolic space. Hyperbolic space is more suitable for processing meshes than Euclidean space, proven in traditional computer graphics. Researchers project meshes into hyperbolic space to learn parameterized representation in shape analysis and model registration tasks~\cite{jin2006computing,shi2013hyperbolic,shi2016shape,aigerman2016hyperbolic}. We embed hand-object meshes into hyperbolic space to learn geometry information. 

\section{Preliminaries}
\subsection{Hyperbolic Space}
Hyperbolic space is a non-Euclidean space that can be represented as a Riemannian manifold with a constant negative curvature. There are multiple isometric hyperbolic models, including  Poincar\'{e} ball model, Lorentz model, and Klein model. In this work, we use the Poincar\'{e} ball model because it is a conformal geometry, \ie geometry-preserving.

The Poincar\'{e} ball model is defined as a Riemannian manifold $\left ( B_{c}^{n},g_{x}^{B}\right )$, where $x$ is a point in a Riemannian manifold and B represents the Poincar\'{e} ball model. $B_{c}^{n}=\left \{x\in \mathbb{R}^{n}:\left \| x\right \|^{2}<-\frac{1}{c}\right \}$ is an n-dimensional ball with radius $\frac{1}{\sqrt{|c|}}$, and $c$ $(c<0)$ is the negative curvature of the ball. $g_{x}^{B}=(\lambda _{x}^{c})^{2}g^{E}$ is its Riemannian metric, which is conformal to the Euclidean metric $g^{E}$ with the conformal factor $\lambda _{x}^{c}=\frac{2}{1-\left \| x\right \|^{2}}$.

To project points from Euclidean space to hyperbolic space, a mapping function called the exponential mapping function $\text{Exp}: \tau _{x}M\rightarrow M$ is defined, as explained in~\cite{yang2022hyperbolic}. Here, $M$ refers to the hyperbolic space in the Poincar\'{e} ball model, and $\tau _{x}M$ is its tangent space. The logarithmic map $\text{Log}$ is the inverse of $\text{Exp}$ and maps points from hyperbolic space to its tangent space, also described in~\cite{yang2022hyperbolic}. If we have two points $x$ and $y$ in $B_{c}^{n}$, the distance between them is the geodesic distance $d(x,y)$. This distance is defined as the shortest length of a curve, as further explained in~\cite{yang2022hyperbolic}.

\begin{figure*}[ht]
\centering
\includegraphics[width=0.95\linewidth]{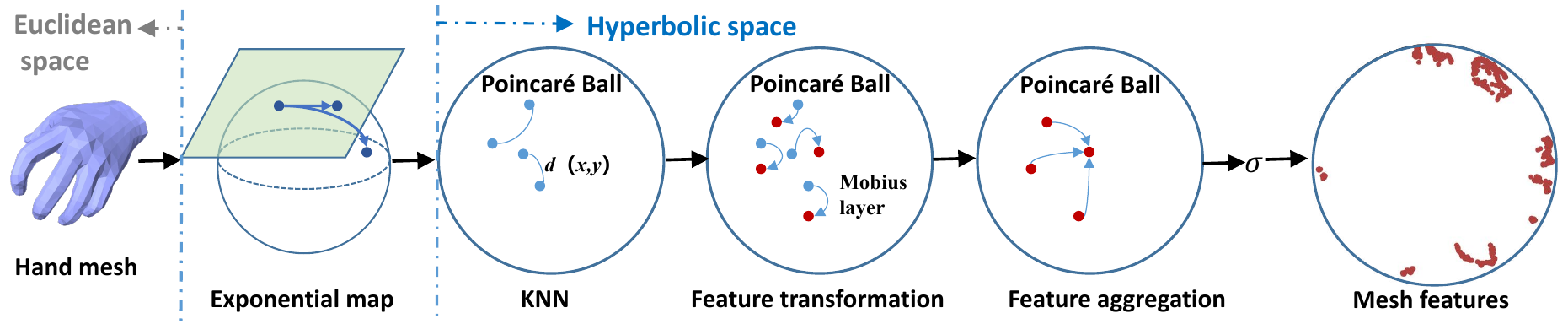}
\caption{This figure illustrates the pipeline of DHGC, which involves several steps.  A given mesh is projected from Euclidean to hyperbolic space using the exponential function. We then conduct dynamic graph construction and employ hyperbolic graph convolution to learn the geometry features of the mesh.}
\label{fig:hgcnone}       
\end{figure*}

\subsection{Hyperbolic Graph Neural Network}
Hyperbolic Graph Neural Networks (HGNN)~\cite{liu2019hyperbolic} generalizes graph neural networks to hyperbolic space. In comparison to graph neural networks in Euclidean space, HGNN is more suitable for tree-like data and therefore learns more powerful geometrical representations~\cite{yang2022hyperbolic}. HGNN consists of four steps: feature projection, feature transformation, neighborhood aggregation, and activation. For the $l$-th layer in HGNN, given a graph $\mathcal{G}=(\mathcal{V},\mathcal{E})$ with a vertex set $\mathcal{V}$ and an edge set $\mathcal{E}$, $x_{i}^{l-1,E}\in \mathcal{V}$ is the input node feature for $i$-th vertex in Euclidean space. The feature projection is to project node features to hyperbolic space by $\text{Exp}$ function. The feature transformation is usually operated by a $M\ddot{o}bius$ layer~\cite{kochurov2020geoopt}, which involves $M\ddot{o}bius$ vector multiplication $\otimes$ and $M\ddot{o}bius$ bias addition $\oplus$. The neighborhood features are aggregated by hyperbolic aggregation functions, $\text{AGG}^{B}$. The last is a non-linear hyperbolic activation, $\sigma ^{B}$. In short, a hyperbolic graph convolution layer can be formulated as:
\begin{gather}
\label{eq:hyperbolic gnn}
    x_{i}^{l-1,B} = \text{Exp}(x_{i}^{l-1,E}), \\
     h_{i}^{l,B} = x_{i}^{l-1,B} \otimes W^{l} \oplus b^{l},\\
     y_{i}^{l,B} = \text{AGG}^{B}(h^{l,B}), \\
     x_{i}^{l,B} = \sigma ^{B}(y_{i}^{l,B}).
\end{gather}%
For more details on the functions, please refer to~\cite{yang2022hyperbolic}.

\section{Methodology}
In this section, we present our novel method for hand object reconstruction, called the Dynamic Hyperbolic Attention Network (DHANet). As shown in Figure \ref{fig:pipeline}, our approach consists of a two-branch network that jointly reconstructs both the hand and object meshes. Specifically, our method comprises three main steps: 1) Image-to-mesh estimation (Section \ref{sec:prerecons}), 2) Dynamic Hyperbolic Graph Convolution for learning mesh features (Section \ref{sec:dhgc}), and 3) Image-attention Hyperbolic Graph Convolution for modeling the hand-object interaction (Section \ref{sec:imageattention}).

\subsection{Image-to-mesh estimation}\label{sec:prerecons}
As depicted in \cref{fig:pipeline}, the image-to-mesh estimation step aims to estimate the initial 3D meshes of the hand and object from a given image. Each branch employs an encoder-decoder architecture, where the encoder consists of two pre-trained ResNet-18~\cite{he2016deep} encoders on ImageNet~\cite{russakovsky2015imagenet}. The decoders output the hand and object meshes respectively.

\textbf{Hand Reconstruction Decoder.} The hand reconstruction decoder predicts the hand parameters from image features using the MANO model~\cite{romero2017embodied}, which is an articulated mesh deformation model rigged with 21 skeleton joints. The MANO model is represented by a differentiable function $D(\beta,\theta)$, where $\theta \in \mathbb {R}^{51}$ denotes the shape parameters and $\beta \in \mathbb{R}^{10}$ denotes the pose parameters. We employ a multi-layer perceptron (MLP) to directly regress $\beta$ and $\theta$ from the image features. Then, a differentiable MANO layer~\cite{hasson2019learning} applies $D$ to generate a hand MANO model from $\beta$ and $\theta$. The hand mesh of the MANO model is defined as $m_{h}=(v_{h},f_{h})$, where $v_{h}\in \mathbb {R}^{778\times3}$ denotes the mesh vertices and $f_{h}\in \mathbb {R}^{1538\times3}$ denotes the mesh faces. The supervision signal for this branch comes from the L2 loss, which consists of the L2 distance between the predicted mesh vertices and the ground truth mesh vertices, as well as the L2 distance between the predicted joint positions and the ground truth joint positions.

\textbf{Object Reconstruction Decoder.} The objective of the decoder for object reconstruction is to predict the 3D object mesh from the image features. We employ AtlasNet~\cite{groueix2018papier} as the object decoder, following the approach of existing methods such as~\cite{hasson2019learning,tse2022collaborative, chen2022alignsdf}. The AtlasNet branch takes the image features from the encoder and generates the object mesh $m_{o}=(v_{o},f_{o})$, where $v_{o}\in \mathbb {R}^{642\times3}$ represents the mesh vertices and $f_{o} \in \mathbb {R}^{1280\times3}$ represents the mesh faces. The branch is trained to minimize the Chamfer distance~\cite{groueix2018papier}, which measures the average minimum distance between points on the predicted mesh and the nearest points on the ground truth mesh.
\begin{figure}[ht]
\centering
\includegraphics[width=\linewidth]{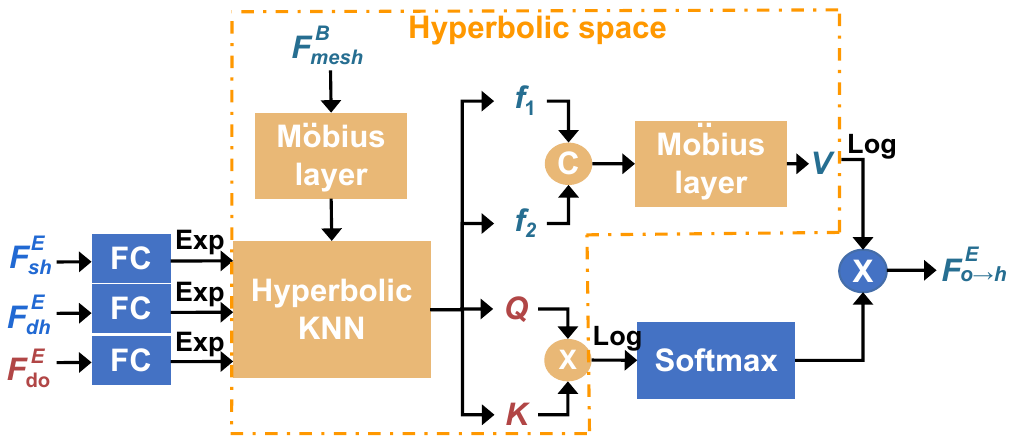}
\caption{Our image attention hyperbolic graph convolution. The operations in the yellow rectangle are implemented in hyperbolic space, while the blues are in Euclidean space.}
\label{fig:hgcntwo}       
\end{figure}
\subsection{Dynamic hyperbolic graph convolution}\label{sec:dhgc}
Hyperbolic space has been shown to be well-suited for processing tree-like graphs due to its exponential growth property, which preserves local and geometric information with low distortion~\cite{peng2021hyperbolic,yang2022hyperbolic}. As meshes are naturally tree-like graphs, we aim to learn mesh features with rich geometric information in hyperbolic space. Inspired by DGCNN~\cite{wang2019dynamic}, which captures the local geometry structure of point clouds in Euclidean space, we propose a dynamic hyperbolic graph convolution to learn mesh features. This module consists of three steps: projection, graph construction, and hyperbolic graph convolution.

\textbf{Projection.} We project the vertices of a mesh $v^{E}\in \mathbb {R}^{n\times3}$ into hyperbolic space using an exponential map function, $v^{B} = \text{Exp} (v^{E})$, as illustrated in \cref{fig:hgcnone}. Here, $v^{B}$ denotes the set of mesh vertices in hyperbolic space.

\textbf{Graph Construction.} To construct a neighborhood graph for the vertices, we employ a hyperbolic k-nearest neighbors (k-NN) algorithm, which searches for the k closest points for each vertex based on the geodesic distance between two vertices, $d(v_{i}^{B},v_{j}^{B})$. This approach allows us to capture the local geometry structure of the mesh in hyperbolic space.

\textbf{Hyperbolic Graph Convolution.} Hyperbolic graph convolution is to learn a neighborhood feature for each vertex, including transforming vertex features on an $m$-dimensional Poincar\'{e} ball by a $M\ddot{o}bius$ layer~\cite{kochurov2020geoopt}, aggregating and activating neighborhood features, as shown in \cref{fig:hgcnone}. This whole process can be formulated as
\begin{equation}
    v^{l,B} = \sigma ^{B} (AGG^{B} (M\ddot{o}bius(\text{exp}(v^{l-1,E})))).
\end{equation} For the aggregation function, we adapt mean aggregation in Poincar\'{e} ball, which returns the Einstein midpoint among vertices in a $k$-neighborhood~\cite{yang2022hyperbolic}. Compared to EdgeConv in DGCNN ~\cite{wang2019dynamic}, DHGC solely focuses on learning pointwise node features without considering edge features, as there is no defined edge vector in hyperbolic space unlike in Euclidean space.

\subsection{Image-attention hyperbolic graph convolution}\label{sec:imageattention}
As mentioned in \cref{sec:related gnn}, due to the exponential growth of distance in hyperbolic space, image features projected to hyperbolic space are more expressive for semantic segmentation~\cite{hyperbolicimage} and image classification~\cite{hyperbolicembedding}. Inspired by these works, we project image features to hyperbolic space. Projected image features preserve the spatial relationship between hand and object, which is beneficial for modeling hand-object interaction, as shown in \cref{fig:teaser}. Hence, we propose an image-attention hyperbolic graph convolution to learn geometry-image multi-modal features, modeling hand-object interaction. As shown in \cref{fig:hgcntwo}, this module consists of four steps, projection, neighborhood graph construction, feature transformation, and image attention. 

\textbf{Inputs.} Taking hand reconstruction as an example, this module takes as input image features in Euclidean space and mesh features in hyperbolic space denoted as $F_{mesh}^{B}$. The image features include shallow image features of the hand $F_{sh}^{E}$, deep image features of the hand $F_{dh}^{E}$, and deep image features of the object $F_{do}^{E}$. To ensure consistency in dimensions, fully connected layers are applied to the image features

\textbf{Projection.} We use the exponential function defined in \cref{eq:hyperbolic gnn} to map the image features to hyperbolic space. This results in obtaining image features in hyperbolic space denoted as $F_{sh}^{B}$, $F_{dh}^{B}$, $F_{do}^{B}$.

\textbf{Graph Construction.} To construct the $k$-neighborhood for each vertex in $F_{mesh}^{B}$, we utilize a hyperbolic KNN algorithm Specifically, We construct four types of $k$-neighborhood for each vertex. These four neighborhoods of each vertex are successively composed of $k$ mesh features, $k$ shallow image features, $k$ deep hand image features and $k$ deep object image features, which are defined as $f_{1}$, $f_{2}$, $Q$ and $K$. Through building four types of $k$-neighborhood, image features, and mesh features are aligned in a unified hyperbolic space.

\textbf{Feature Transformation}. Mesh features are enhanced by similar shallow image features. In a neighborhood, mesh features $f_{1}$ are concatenated with similar shallow image features $f_{2}$. The concatenated feature is transformed into $V$ with a similar dimension as $Q$ and $K$, by a $M\ddot{o}bius$ layer, formulated as:
\begin{equation}
    V = M\ddot{o}bius(Cat(f_{1},f_{2})),
\end{equation} where $Cat$ represents the concatenation operation. 

\textbf{Image Attention.} We define the image attention to model the hand-object interaction. $V$ indicates hand mesh features. $Q$ refers to deep image features of hands, which are similar to hand mesh, while $K$ refers to deep image features  of objects, which are similar to hand mesh. Then we use the object image feature to fetch the hand image feature and hand mesh feature, as shown \cref{fig:hgcntwo}. The process can be formulated as 
\begin{equation}
    F_{o\rightarrow h}^{E} = \text{softmax}(\frac{\text{Log}(Q)\text{Log}(K)^{T}}{\sqrt{d}})\text{Log}(V),
\end{equation}
where $F_{o\rightarrow h}^{E}$ is the hand-object attention mesh features encoding the interaction between hand and object, and $d$ is a normalization constant. For ease of calculation, we map features by Log function into Euclidean space. At last, image-attention hyperbolic graph convolution learns geometry-image multi-modal features, concatenated with image features from encoders to reconstruct a mesh by decoders, as shown in \cref{fig:pipeline}.

\section{Implement Details of DHANet}
\textbf{Architecture.} Given an input image $I$ with size $256\times 256$, DHANet reconstructs a hand mesh $m_{h}$ of size $778\times 3$, and an object mesh $m_{o}$ of size $642\times 3$. Our DHANet is a two-branch network, one for hand reconstruction and the other one for object reconstruction. In the hand branch, the encoder, ResNet-18~\cite{resnet}, extracts 512-dimensional image features $F_{h}$, shallow image features $F_{sh}^{E}$ with size $64\times 64\times 32$, and deep image features $F_{dh}^{E}$ with size $8\times 8\times 64$. Then the decoder consisted of fully connected layers and a MANO layer initially reconstructs a hand mesh $m_{h}$ from $F_{h}$. With the hand mesh as input, two stacked DHGC layers learn mesh features $F_{h,mesh}^{B}$ of size $778\times 64$ in hyperbolic space. The IHGC learns enhanced 32-dimensional mesh features, $F_{o\rightarrow h}^{E}$, inputting $F_{h,mesh}^{B}$, $F_{sh}^{E}$, $F_{dh}^{E}$ and deep image features of objects $F_{do}^{E}$. At last, mesh features are concatenated with 512-dimensional image features, reconstructing hand and object mesh by the decoder. The architecture of the object branch is similar to the hand branch. In order to reconstruct the photo-consistent hand and object, we also adopt two fully connected layers to estimate the 3D offset coordinates $T$ for the translation and a scalar $S$ for the scale, as in ~\cite{hasson2019learning}. For more details on the detailed architecture of our DHANet please refer to the supplementary material.

\textbf{Loss function.} To supervise the training of our DHANet, we use the loss items in Hasson \etal.~\cite{hasson2019learning}, except for the contact loss term, defined as:
\begin{gather}
\label{equ:loss}
    \mathcal{L}=\mathcal{L}_{hand}+\mathcal{L}_{obj} \\
    \mathcal{L}_{hand}=\mathcal{L}_{V_{hand}}+\mathcal{L}_{J}+\mathcal{L}_{\beta} \\
    \mathcal{L}_{obj}=\mathcal{L}_{V_{obj}}+\mathcal{L}_{T}+\mathcal{L}_{S}.
\end{gather}
In \cref{equ:loss}, $\mathcal{L}_{hand}$ for the hand branch consists of the L2 loss of vertex positions $\mathcal{L}_{V_{hand}}$, the L2 loss of hand joints $\mathcal{L}_{J}$ and the L2 loss of the hand shape $\mathcal{L}_{\beta}$. $\mathcal{L}_{obj}$ for the object branch includes the Chamefer distance $\mathcal{L}_{V_{obj}}$, the L2 loss of object scale $\mathcal{L}_{S}$ and the L2 loss of object translation $\mathcal{L}_{T}$. $\mathcal{L}_{T}$ is defined as $\mathcal{L}_{T}=\left \| T-\hat{T}\right \|_{2}^{2}$, where $\hat{T}$ is the ground truth object centroid in hand-relative coordinates. $\mathcal{L}_{S}$ is defined as $\mathcal{L}_{S}=\left \| S-\hat{S}\right \|_{2}^{2}$, where $\hat{S}$ is the ground truth maximum radius of the centroid-centered object.

\section{Experiments}
\subsection{Datasets}

\textbf{Obman} is a large-scale synthetic image dataset of hands grasping objects~\cite{hasson2019learning}. The objects in Obman are 8 types of common items, whose models are selected from the ShapeNet~\cite{chang2015shapenet} dataset. The hands in this dataset are modeled with MANO~\cite{MANO:SIGGRAPHASIA:2017}. The dataset is labeled with 3D hand and object meshes, divided into 141K training frames and 6K test frames.

\textbf{First-person hand benchmark (FHB)} is a real egocentric RGB-D videos dataset about hand-object interaction~\cite{FHB}. There are 105,459 RGB-D frames annotated with 3D object meshes for 4 items and the 3D location of hand joints. We use the same way to divide a training set and a testing set, like~\cite{hasson2019learning}. To be consistent with the existing methods~\cite{hasson2019learning, tse2022collaborative}, we exclude the milk model and filter frames in which the hand is further than 10 mm from the object. This subset of FHB is called $\text{FHB}^{-}$.

\textbf{HO-3D} is also a real image dataset for hand-object interaction~\cite{hampali2020honnotate}. The objects in HO-3D are 10 objects from YCB dataset~\cite{Ycb}. The dataset contains hand-object 3D pose annotated RGB images and their corresponding depth maps. Our experiment uses HO-3D (version 2) split into 70K training images and 10K evaluation images as in~\cite{hasson2020leveraging}.

\begin{figure}[tb]
\centering
\includegraphics[width=0.9\linewidth]{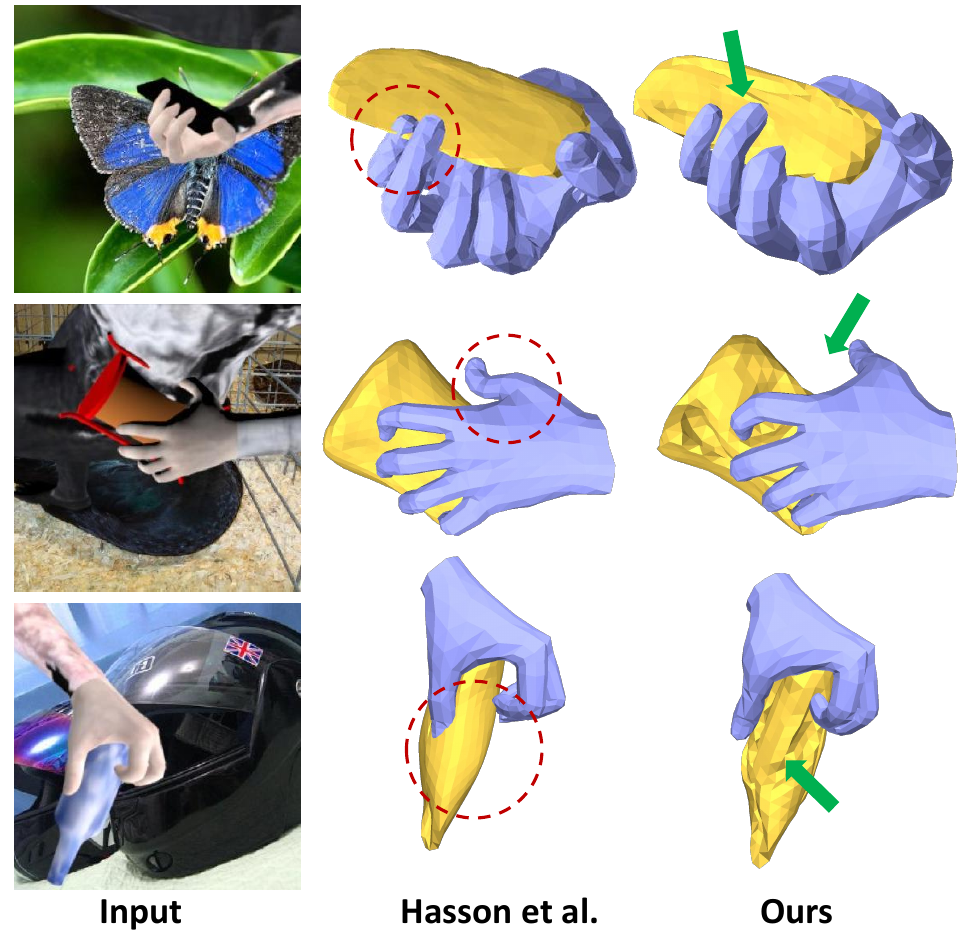}
\caption{Qualitative comparison with Hasson \etal.~\cite{hasson2019learning} on Obman dataset~\cite{hasson2019learning}. The red circles highlight the errors from Hasson \etal.~\cite{hasson2019learning}. The green arrows point to improvements of our method.}
\label{fig:sota_obman}       
\end{figure}

\subsection{Evaluation metrics}
The reconstruction quality of hands and objects are evaluated with the following metrics.

\textbf{Hand error.} The mean end-point error (mm) over 21 joints and the mean vertices error of meshes are computed to evaluate the hand reconstruction. 

\textbf{Object error.} To evaluate the object reconstruction, we report the Chamfer distance (mm) between points sampled on the ground truth mesh and vertices of the predicted mesh.

\textbf{Contact metrics.} Reconstructed hand-object should be impenetrable according to the laws of physics. For assessing the physical validity of the results, we also adopt penetration depth (mm) and intersection volume ($cm^{3}$) as~\cite{hasson2019learning, tse2022collaborative}. Penetration depth is the maximum distance between hand mesh and object mesh when the hand collides with the object. Otherwise, the penetration depth is 0. Intersection volume is the volume of the interaction area btween the hand and object. We compute the volume by voxelizing the hand and object under a voxel size of 0.5 cm.

\subsection{Training details} 
We adopt the work of Hasson \etal.~\cite{hasson2019learning} as the backbone to pre-estimate rough hand-object meshes, namely baseline. The model parameters of the baseline are initialized by the pre-trained model of Hasson \etal.~\cite{hasson2019learning}. We choose the Riemannian Adam optimizer to train our DHANet, since ~\cite{kochurov2020geoopt} verified that the Riemmanian Adam optimizer speeds up model convergence for hyperbolic space and works as the standard Adam optimizer for Euclidean space. The training for different datasets is different. For the Obman dataset, the training strategy is the same as~\cite{hasson2019learning}. We first train the object branch for 100 epochs at a learning rate $10^{-4}$, then train the hand branch for 100 epochs at a learning rate $10^{-4}$ while freezing the object branch. For datasets of real scenes, HO-3d and $\text{FHB}^{-}$, we train the hand and object branches together for 300 epochs with a learning rate of $10^{-4}$, then train them for other 300 epochs with a learning rate of $10^{-5}$.

\begin{table}[tp]
\begin{center}
\resizebox{\linewidth}{!}{
\begin{tabular}{l|cccc}
\hline
Methods       & \begin{tabular}[c]{@{}c@{}}Hand \\ error\end{tabular} & \begin{tabular}[c]{@{}c@{}}Object \\ error\end{tabular} & \begin{tabular}[c]{@{}c@{}}Max. \\ penetra.\end{tabular} & \begin{tabular}[c]{@{}c@{}}Intersect. \\ vol.\end{tabular} \\ \hline
Hasson \etal.~\cite{hasson2019learning}  & 11.6  & 641.5    & 9.5    & 12.3    \\
Tse \etal.~\cite{tse2022collaborative}  & \textbf{9.1}   & \textbf{385.7}    & \textbf{7.4}    & \textbf{9.3}        \\ \hline
Ours                                     & 10.2  & 529.3    &  9.3     &    10.4   \\ \hline
\end{tabular}}
\end{center}
\caption{Comparison to state-of-the-art methods on Obman dataset~\cite{hasson2019learning}. The hand error is calculated on joints. Here we report the result of Hasson \etal.~\cite{hasson2019learning} without contact loss. ``Max penetration" is shortened to ``Max. penetra.". ``Intersection volume" is shortened to ``Intersect. vol." }
\label{tab:obman sota}
\end{table}
\subsection{Hand-object reconstruction results}
\textbf{Method for comparison.} In the single image hand-object reconstruction field, there are a few methods~\cite{hasson2019learning,cao2021reconstructing,yang2021cpf,tse2022collaborative,li2021artiboost}, which represent a hand as MANO model and represent an object as 3D mesh. While existing methods include two categories, one for known object models~\cite{yang2021cpf,cao2021reconstructing,li2021artiboost}, the other for unknown object models~\cite{hasson2019learning,tse2022collaborative}, Our method belongs to the latter category. There are still some hand-object reconstruction works based on SDF~\cite{chen2022alignsdf,ye2022s}, representing objects as a dense 3D mesh, while in our work we reconstruct an object as a simple mesh with 642 vertices. Hence, we compare our method with~\cite{hasson2019learning} and~\cite{tse2022collaborative}.  

\begin{table}[tb]
\begin{center}
\resizebox{\linewidth}{!}{
\begin{tabular}{l|cccc}
\hline
Methods       & \begin{tabular}[c]{@{}c@{}}Hand \\ error\end{tabular} & \begin{tabular}[c]{@{}c@{}}Object \\ error\end{tabular} & \begin{tabular}[c]{@{}c@{}}Max. \\ penetra.\end{tabular} & \begin{tabular}[c]{@{}c@{}}Intersect. \\ vol.\end{tabular} \\ \hline
Hasson \etal.~\cite{hasson2019learning}  & 28.1  & 1579.2    & 18.7    & 26.9    \\
Tse \etal.~\cite{tse2022collaborative}  & 25.3   & 1445.0    & 16.1    & \textbf{14.7}        \\ \hline
Ours                                     & \textbf{23.8}  & \textbf{1236.0}    &  \textbf{14.43}  & 20.7  \\ \hline
\end{tabular}}
\end{center}
\caption{Comparison to state-of-the-art methods on $\text{FHB}^{-}$ dataset~\cite{FHB}. The hand error is calculated on joints. Here we report the result of Hasson \etal.~\cite{hasson2019learning} without contact loss. ``Max penetration" is shortened to ``Max. penetra.". ``Intersection volume" is shortened to ``Intersect. vol."}
\label{tab:FHB sota}
\end{table}

\begin{table}[tb]
\begin{center}
\begin{tabular}{l|cc}
\hline
Methods       & \begin{tabular}[c]{@{}c@{}}Hand \\ error\end{tabular} & \begin{tabular}[c]{@{}c@{}}Object \\ error\end{tabular}  \\ \hline
Hasson \etal.~\cite{hasson2020leveraging}  & 14.7  & 26.8     \\
Cao \etal.~\cite{cao2021reconstructing}  &  9.7      &  19.9        \\
Tse \etal.~\cite{tse2022collaborative}  & 10.9   &  -        \\ \hline
Ours                                     & \textbf{6.1}  & \textbf{13.8}    \\ \hline
\end{tabular}
\end{center}
\caption{Comparison to state-of-the-art methods on HO-3d dataset~\cite{hampali2020honnotate}. The hand error is calculated on the vertices of the hand mesh.}
\label{tab:HO-3d sota}
\end{table}

\textbf{Results.} \cref{tab:obman sota} indicates our method achieves better results on Obman dataset~\cite{hasson2019learning} than the baseline method~\cite{hasson2019learning} in hand and object errors. Compared with the baseline method~\cite{hasson2019learning}, our method yields a smaller hand error of 10.7 mm vs. 11.6 mm and a smaller object error of 563.5 mm vs. 641.5 mm. Our method also achieves better results on contact metrics. As shown in \cref{fig:sota_obman}, our method reconstructs better the fine-grained pose and shape of hands with respect to the input image. Like the drum in \cref{fig:sota_obman}, the reconstructed drum by our method is more consistent with the original shape in the image. And it can be observed that hands reconstructed by our method are penetrated less with objects. This suggests that our method better models hand-object interaction. However, the performance of our method is less than Tse \etal.~\cite{tse2022collaborative} in \cref{tab:obman sota}. The reason is that the work of Tse \etal.~\cite{tse2022collaborative} is a dual-iterative network, in contrast to our DHANet that operates without iteration. While the two iterations of Tse \etal.~\cite{tse2022collaborative} yield a good result,  they increase the model parameter simultaneously.

\begin{figure}[tb]
\centering
\includegraphics[width=0.9\linewidth]{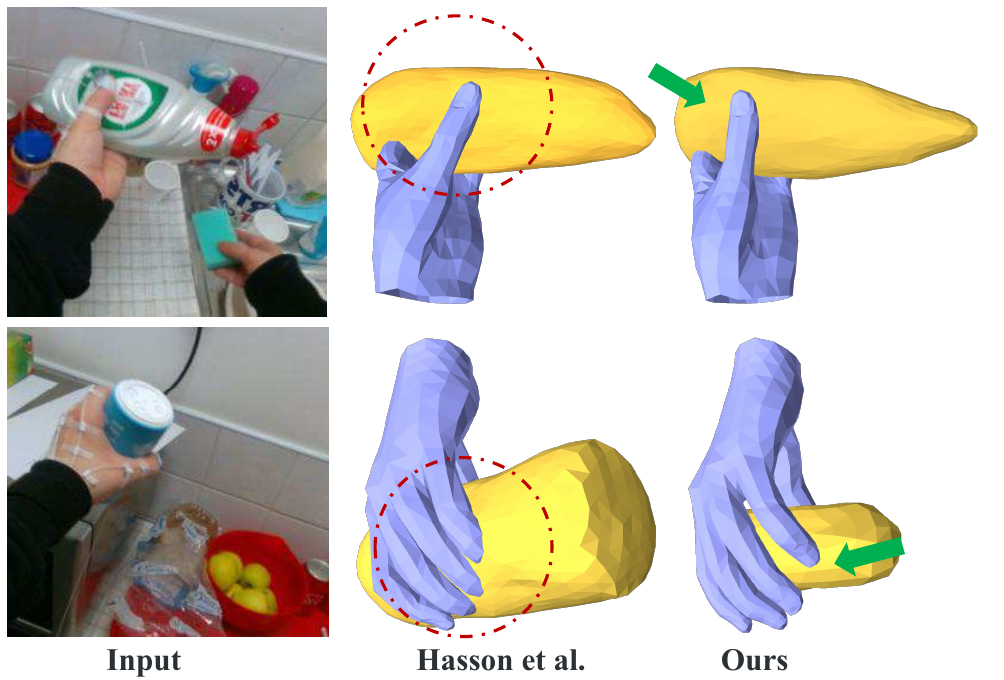}
\caption{Qualitative comparison with Hasson \etal.~\cite{hasson2019learning} on $\text{FHB}^{-}$ dataset~\cite{FHB}.The red circles hightlight the errors from Hasson \etal.~\cite{hasson2019learning}. The green arrows point to improvements of our method. }
\label{fig:sota_fhb}       
\end{figure}

The experimental results compared with existing methods on $\text{FHB}^{-}$ dataset~\cite{FHB} is listed in \cref{tab:FHB sota}. In $\text{FHB}^{-}$ dataset, our method achieves SOTA results whit smaller hand error (23.8 mm) and smaller object error (1236.0 mm). The qualitative results of this dataset are shown in \cref{fig:sota_fhb}. Our method exceeds the work of Tse \etal.~\cite{tse2022collaborative} on $\text{FHB}^{-}$ dataset in \cref{tab:FHB sota}, but slightly worse on Obman dataset in \cref{tab:obman sota}. The reason for this result is dataset difference: Obman dataset is a synthetic dataset with complex and various backgrounds, while $\text{FHB}^{-}$ dataset is a real-world dataset with a simplistic kitchen environment background. This results in shallow image features learned on Obman containing more irrelevant background information. The more irrelevant features are, the more loose their projections are in hyperbolic space, and vice versa. We drew the visualization of it in the supplemental material. Those loose shallow image features in hyperbolic space are unfavorable to searching the neighborhood between those features and mesh features. This causes our approach to work slightly worse in Obman dataset.

And the comparison results on HO-3d~\cite{hampali2020honnotate} are shown in \cref{tab:HO-3d sota}. We also reach SOTA results on the hand error and the object error. 
$\text{FHB}^{-}$ dataset and HO-3d are captured in real scenes, not synthetic data. The decent results manifest our method can handle not only synthetic data but also real-world cases.

\begin{figure*}[ht]
\centering
\includegraphics[width=0.95\linewidth]{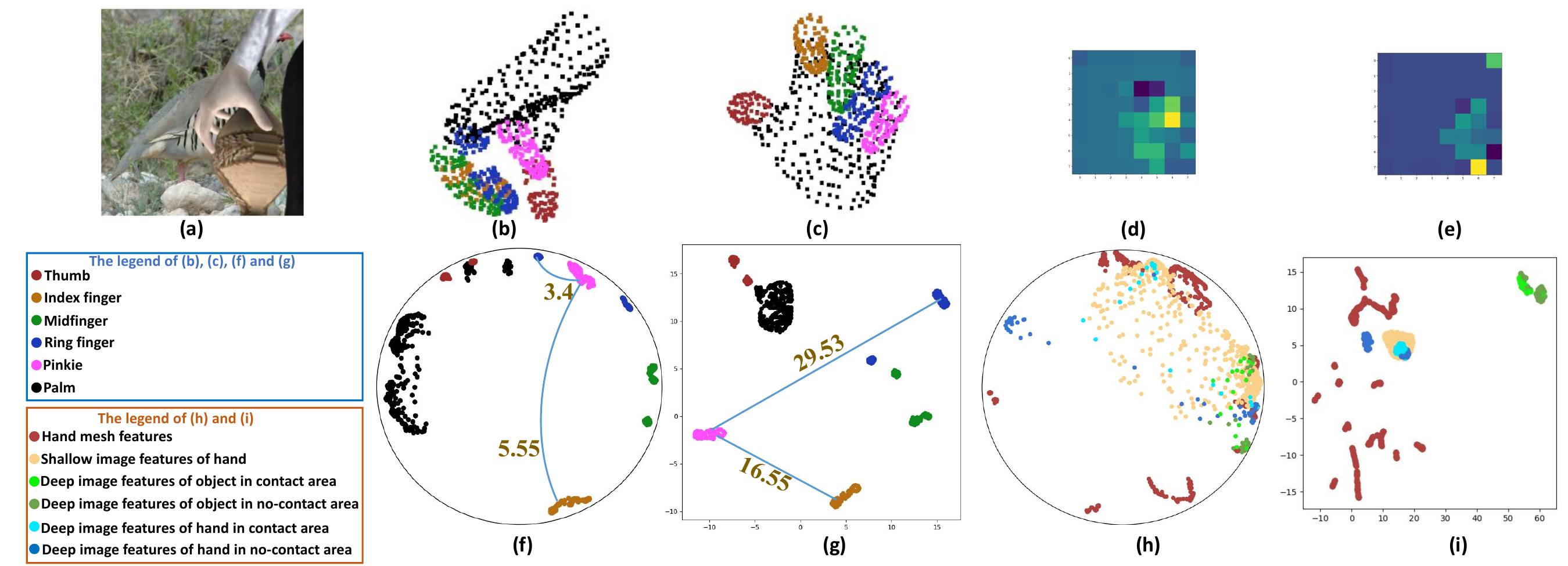}
\caption{Visualization of features in hyperbolic space and Euclidean space. (a): a sample image from Obman dataset~\cite{hasson2019learning}. (b): vertices of the hand mesh reconstructed from (a). (c) is rotated by (b). (d): the hand deep image features from the encoder of the hand branch. (e): the object deep image features from the encoder of the object branch. The description of (f), (g), (h), (i), and (j) is in \cref{sec: vis hyper}.}
\label{fig:vis_mesh}       
\end{figure*}

\subsection{Ablation study}
We conducted ablation studies to demonstrate the effectiveness of our proposed dynamic hyperbolic graph convolution (DHGC) and image-attention hyperbolic graph convolution (IHGC). As shown in \cref{tab:abliation}, adding DHGC with baseline reduces the hand error to 10.9 mm while reducing the object error to 582.9 mm. This suggests that the mesh feature learned by DHGC provides richer geometric information. Furthermore, IHGC further improves the reconstruction results, which further reduced the hand and object error to 10.2 mm and 529.3 mm. And the performance on contact metrics also declined. These results demonstrate that IHGC effectively enhances mesh features with image features while modeling hand-object interactions.

In order to verify the superiority of our method in hyperbolic space, we also implement dynamic graph convolution and image-attention graph convolution in Euclidean space. The comparison results are enumerated in \cref{tab:abliation}. We can observe that these two modules in Euclidean space have improved from the baseline~\cite{hasson2019learning}, while the improvement is less than ours in hyperbolic space. It proves quantitatively that our method achieves better performance in hyperbolic space than in Euclidean space.

\begin{table}[htbp]
\begin{center}
\resizebox{\linewidth}{!}{
\begin{tabular}{l|cccc}
\hline
Methods       & \begin{tabular}[c]{@{}c@{}}Hand \\ error\end{tabular} & \begin{tabular}[c]{@{}c@{}}Object \\ error\end{tabular} & \begin{tabular}[c]{@{}c@{}}Max. \\ penetra.\end{tabular} & \begin{tabular}[c]{@{}c@{}}Intersect. \\ vol.\end{tabular} \\ \hline
Baseline~\cite{hasson2019learning}  & 11.6  & 641.5    & 9.5    & 12.3    \\ \hline
Baseline+1(EU) & 11.6   & 589.2   & 10.8    & 13.5        \\ 
Baseline+1+2(EU) & 11.1  & 586.1  & 11.2    & 10.7       \\ \hline
Baseline+1(H)    & 10.9  & 582.9  & 10.7    & 10.9      \\   
Baseline+1+2(H)  & \textbf{10.2}  & \textbf{529.3}    &  \textbf{9.3}     &  \textbf{10.4}   \\ \hline
\end{tabular}}
\end{center}
\caption{Ablations on modules and feature spaces. 1 refers to dynamic hyperbolic graph convolution. 2 refers to image-attention hyperbolic graph convolution. EU represents the operation in Euclidean space, while H represents hyperbolic space.}
\label{tab:abliation}
\end{table}

\subsection{Visual analysis of hyperbolic learning}
\label{sec: vis hyper}
We further prove the superiority of our method by visual analysis of features in hyperbolic space, as shown in \cref{fig:vis_mesh}. To facilitate observation, we use UMAP~\cite{umap} to project features to 2 dimensions, as in~\cite{montanaro2022rethinking}. Given an image as in \cref{fig:vis_mesh} (a), we visualized the distribution of the corresponding 3D mesh in hyperbolic space and Euclidean space, depicted in \cref{fig:vis_mesh} (f) and \cref{fig:vis_mesh} (g). As shown in \cref{fig:vis_mesh} (c), pink points are near blue points and far from brown points. The relative position is reflected in hyperbolic space, as depicted in \cref{fig:vis_mesh} (f), but not in Euclidean space, as depicted in (g). It indicates that embedding mesh into hyperbolic space can preserve the geometry properties of the mesh.

In image-attention hyperbolic graph convolution, we project mesh features, shallow image features, deep image features of hand, and deep image features of object to hyperbolic space. In \cref{fig:vis_mesh} (h), some yellow points are overlapping with red points. It indicates that shallow image features are aligned with mesh features in hyperbolic space. However, shallow image features and mesh features are separated in Euclidean space, as shown in \cref{fig:vis_mesh} (i). Aligned features are conducive to feature learning in a unified space. In addition, there are overlapping regions in deep image features of hand and object, as shown in \cref{fig:vis_mesh} (d) and \cref{fig:vis_mesh} (e), expressing the area of hand-object interaction. It is reflected in hyperbolic space, as shown in \cref{fig:vis_mesh} (h). Some light blue points are close to a few light green points, others vice versa. The closer region in hyperbolic space represents the area of hand-object interaction in the image. Furthermore, the spatial relationship is not expressed in Euclidean space. As shown in \cref{fig:vis_mesh} (i), the light blue points are far from the light green points. This highlights the ability of hyperbolic space to align multi-modal features and preserve spatial relationships.


\section{Conclusion}
In this work, we propose a dynamic hyperbolic graph neural network (\methodname) for hand object reconstruction. Our method applies hyperbolic neural networks for the first time in this task. By leveraging hyperbolic space, we design a dynamic hyperbolic graph convolution that captures rich geometry information in mesh features. By projecting multi-modal features to a unified hyperbolic space, we define a more accurate representation of geometry-image features. To model the hand-object interaction, we introduce an attention-based hyperbolic graph convolution that enhances mesh features with image features. Our method outperforms state-of-the-art methods on public datasets, achieving more accurate reconstruction of hand and object meshes. This approach offers a new perspective for hand object reconstruction in hyperbolic space, with promising opportunities for future research. As for future work, since image features are vital for modeling interaction and affect our performance, we plan to explore compositional image features, such as extracted from hand or object parts.

\section*{Acknowledgment}
This work was supported by the National Nature Science Foundation of China under Grant 62272019, in part by China Scholarship Council.

{\small
\bibliographystyle{ieee_fullname}
\bibliography{egbib}
}

\end{document}